\documentclass[letterpaper]{article} 
\usepackage{aaai25}  
\usepackage{times}  
\usepackage{helvet}  
\usepackage{courier}  
\usepackage[hyphens]{url}  
\usepackage{graphicx} 
\usepackage{natbib}  
\usepackage{caption} 
\usepackage{algorithm}
\usepackage{algorithmic}
\usepackage{amsmath}
\usepackage{amsfonts} 
\usepackage{subcaption}
\usepackage{multirow}
\usepackage{booktabs}

\usepackage{newfloat}
\usepackage{listings}
\DeclareCaptionStyle{ruled}{labelfont=normalfont,labelsep=colon,strut=off} 
\floatstyle{ruled}
\newfloat{listing}{tb}{lst}{}
\floatname{listing}{Listing}
\title{HomeDiffusion: Zero-Shot Object Customization with Multi-View Representation Learning for Indoor Scenes}
\author{
    Guoqiu Li,
    Jin Song,
    Yiyun Fei
}
\affiliations{
    Alibaba Group\\
    \{liguoqiu.lgq, songjin.song, yunhun.fyy\}@alibaba-inc.com
}

\usepackage{bibentry}

\begin{document}

\maketitle

\begin{abstract}
Recently, zero-shot object customization generation methods have rapidly developed and shown tremendous potential for applications. For instance, in the e-commerce domain, consumers can observe the visual effect of furniture placed within their personal living spaces or clothes worn on their own bodies.
Many existing approaches perform object customization generation based on diffusion models and extracted  reference object features. However, the generated object significantly diverges from the original reference object in details such as patterns and curves. Particularly for asymmetrical reference objects, the absence of comprehensive multi-viewpoint information prevents the generation of object poses that harmonize with the background scene.
To address these shortcomings, we have constructed a novel dataset comprising multi-angle images of  furniture and indoor scenes. Based on diffusion models, we introduce HomeDiffusion, which can leverage multi-viewpoint images of the same reference object to accurately generate visually harmonious object poses within specified areas of the background scene.
During the diffusion process, we further extract high-fidelity details of the reference object and perform cross-attention with the noise latents in the latent space, thereby ensuring the preservation of details in the customized object generation.
Extensive qualitative and quantitative experiments demonstrate that our method achieves superior performance over other existing zero-shot as well as few-shot object customization approaches.
\end{abstract}

%

\section{Introduction}
\label{sec:intro}
The goal of object customization in image editing is to seamlessly integrate objects from reference images into a specific area of the target edited image, ensuring that they harmoniously align with the background in layout, lighting, perspective, and spatial relationships, while maintaining the object's identity and intrinsic properties, including texture, shape, and distinctive features.
Object customization image editing method opens up a world of possibilities, such as the ability to visualize how a new sofa looks in your living room with just a few clicks, without ever stepping into a store.
The field's potential makes it a significant research interest.

Many state-of-the-art methods \cite{ruiz2023dreambooth,paintbyexample,anydoor} utilize large pre-trained text-to-image models such as Stable Diffusion \cite{rombach2021highresolution} as their backbone networks, leveraging the rich prior knowledge of object categories contained within.
The customized objects typically harmonize with the background in terms of lighting and color tone, while still retaining a considerable amount of their identity features.
However, two key challenges remain highly formidable. 
The first challenge is the acquisition of multi-view object representations.
While various approaches attempt to blend the reference object into an edited image by subtly adjusting the edges' tone and texture for coherence, these methods can sometimes struggle with perspective harmony.
This can result in a mismatch between the object and its environment, leading to a visually awkward appearance.
Particularly for asymmetric objects, it becomes impossible to generate an object with a completely different perspective using a single-view object representation when the orientation of the background is entirely dissimilar.
The second challenge is to preserve the high-fidelity object details.
Although some methods \cite{paintbyexample, anydoor} utilize powerful pre-trained multi-modal or self-supervised image encoders to encode the identity features of the target, ensuring consistency throughout generation. The lack of maintenance of detail information results in generated images that do not meet expectations (refer to Figure \ref{fig:4}).

To address the above challenges, we propose a novel zero-shot object customization image editing framework for indoor scenes, named HomeDiffusion. HomeDiffusion adeptly focuses on embedding high-fidelity objects into edited images using reference images of objects from various viewpoints, ensuring the generated object's perspective is aligned with the background scene. 
Compared to outdoor scenes, indoor scenes have a more defined spatial structure. The placement of objects follows certain patterns and knowledge, making it an ideal and essential scenario for object customization capabilities.
Specifically, we build a multi-view indoor dataset, which comprises a collection of indoor scene images and furniture images captured from multiple viewpoints.
We propose a powerful HD visual encoder that can extract high-resolution details of objects from both global and local perspectives.
Then, we introduce the Multi-view Object Representation Learning (MORL) process, wherein multiple viewpoint images of a furniture item serve as input, and the model is trained to predict another viewpoint of this furniture. Through an extensive self-generated training process utilizing images of objects from multiple viewpoints, the model acquires the capability to learn multi-view representations of the objects.

We further propose the Background-driven Object Customization Learning (BOCL) process.
After the MORL stage, we preserve the model's ability to extract multi-view object representations.
During the BOCL stage, the model is trained to insert the reference object into a designated area of the background image, ensuring that the object's perspective is in harmony with the surrounding scene.
To maintain the object's high-fidelity details, we create a composited image that fuses the reference object with scene information. This image, containing high-fidelity object details, will be used to guide the image generation.
Given the perspective deviation between the reference object in the composite image and the target in the generated image, we further introduce pixel-aligned cross-attention calculations within the latent space, thereby more effectively extracting high-fidelity details from the composite image. Our main contributions are: 
\begin{itemize} 
\item we propose HomeDiffusion, a novel zero-shot object customization method for indoor scenes, which ensures high-fidelity object detail and harmonious perspective blending with the background scene.
\item We propose an HD visual encoder that can extract fine details of objects from both global and local views.
\item we present a MORL process that learns robust multi-view object representation through self-generative training using images from multiple viewpoints.
\item we introduce a BOCL method that harmoniously integrates reference objects into designated areas within background scenes. Furthermore, by executing pixel-aligned cross-attention calculations in the latent space, we effectively preserve the high-fidelity object details.
\item we carry out experiments on our collected ZOC-Indoor-Eval benchmark and the publicly Viton-HD benchmark. The qualitative and quantitative results show that our approach achieves superior performance compared to many zero-shot or few-shot object customization methods.
\end{itemize}

\begin{figure}[t]
    \centering
    \includegraphics[width=0.95\columnwidth]{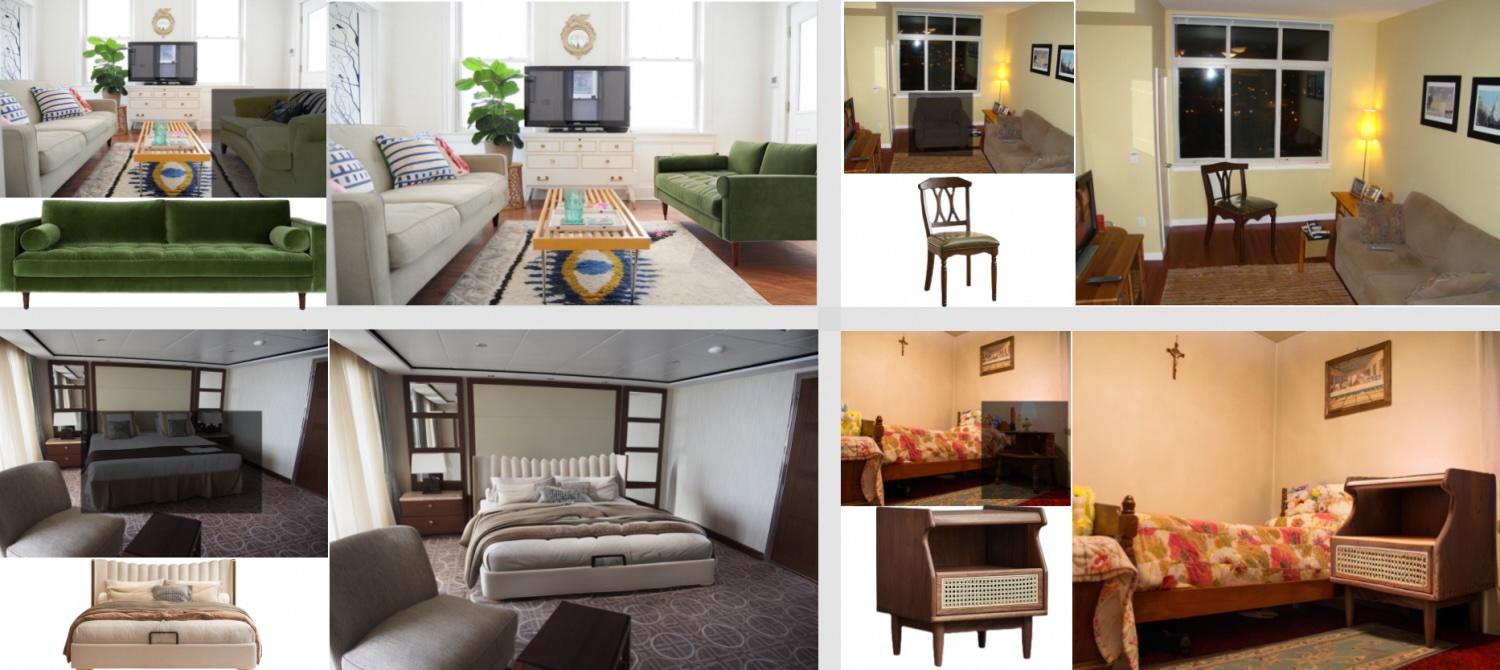}
    \caption{HomeDiffusion enables users to virtually place e-commerce furniture in any scene, ensuring a high degree of fidelity and harmonious perspective with the background. }
    \label{fig:00}
\end{figure}

\section{Related Works}
\paragraph{Diffusion Based Text-to-Image Generation.}
Recently, Diffusion models have achieved significant success in the text-to-image generation field \cite{rombach2021highresolution,2021GLIDE,Imagen,podell2023sdxl,Dai2023EmuEI,balaji2022ediffi,ho2022cascaded,kumari2023ablating}, yielding better results compared to methods such as GAN \cite{DiffuisonBeatGAN,goodfellow2020generative}. Diffusion models create images by learning to progressively recover images from pure Gaussian noise \cite{ho2020denoising}.
Stable Diffusion \cite{rombach2021highresolution} employs a VAE to compress the image from pixel space to latent space, reducing the computational power required for image generation. 
However, in T2I models, it is not possible to implement object embedding through textual guidance.
\paragraph{Diffusion Based Image Editing.}
 There has been some impressive work in image editing using diffusion models \cite{repaint,sdedit,kawar2023imagic,kim2022diffusionclip,prompt2prompt,zhang2023sine}. 
These methods typically use text to guide specific areas of images. However, it's clear that text descriptions alone are not enough to capture all the small, important details needed to describe objects accurately. To solve this problem, some methods use photos of the object as guidance, combining text input to generate images. 
A standout example is Paint by Example \cite{paintbyexample}, the method utilizes the CLIP model \cite{radford2021learning} to extract image embeddings as the image condition injected into the inpainting process; however, the features extracted based on CLIP are not sufficient for preserving the fine details of objects. 
Even with these approach, preserving all details of the object in the final edited image remains a challenging task.

\paragraph{Subject-Driven Image Generation.}
The goal of personalized image generation is to retain as much detail of the object as possible in the generated images, given one or more photos of the object. Recently, several approaches based on pre-trained text-to-image models have been proposed \cite{ruiz2023dreambooth, worthoneword, xiao2023fastcomposer,li2023photomaker,voynov2023p+,shi2023instantbooth,chen2022reimage,CustomDiffusion,SuTI}. DreamBooth \cite{ruiz2023dreambooth} finetunes a pre-trained text-to-image model by training a LoRA model, which works by associating a special word in the prompt with the example images.
Works closely related to ours are AnyDoor \cite{anydoor}. AnyDoor \cite{anydoor}, utilizing DINO-V2 \cite{dinov2} for image feature extraction, is trained a single time and seamlessly generalizes to various object-scene combinations at inference, producing images that richly preserve object details.
However, these methods struggle with achieving high similarity in detail when generating images from multiple viewpoints. Conversely, our method can deliver consistency across various angles and high-fidelity details of objects without the need for test-time training.

\begin{figure*}[t]
  \centering
  \begin{subfigure}{1\columnwidth}
  \centering
    \includegraphics[width=1\columnwidth]{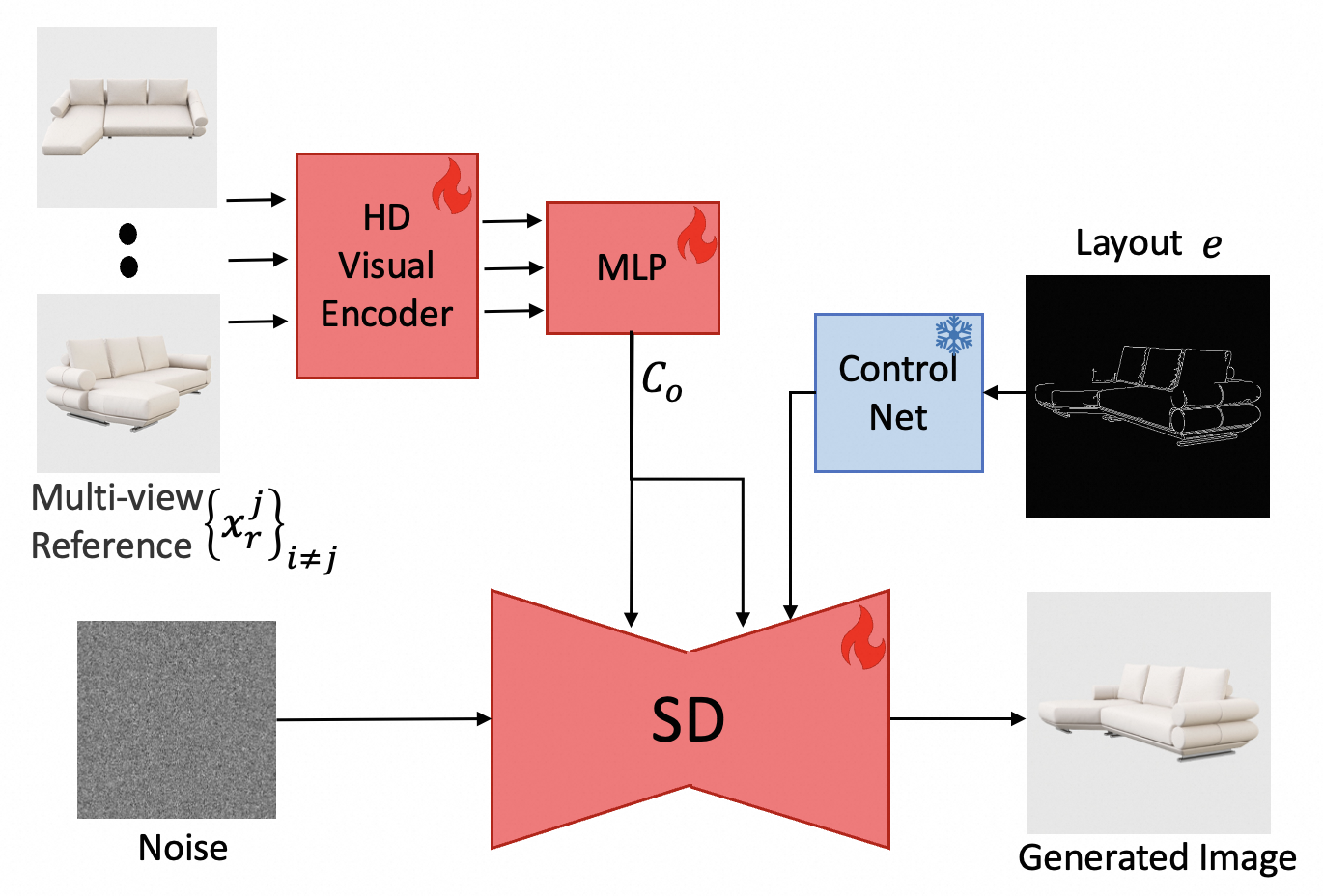}
    \caption{Stage 1: Multi-view Object Representation Learning.}
    \label{fig:1-a}
  \end{subfigure}
  \begin{subfigure}{1\columnwidth}
  \centering
    \includegraphics[width=1\columnwidth]{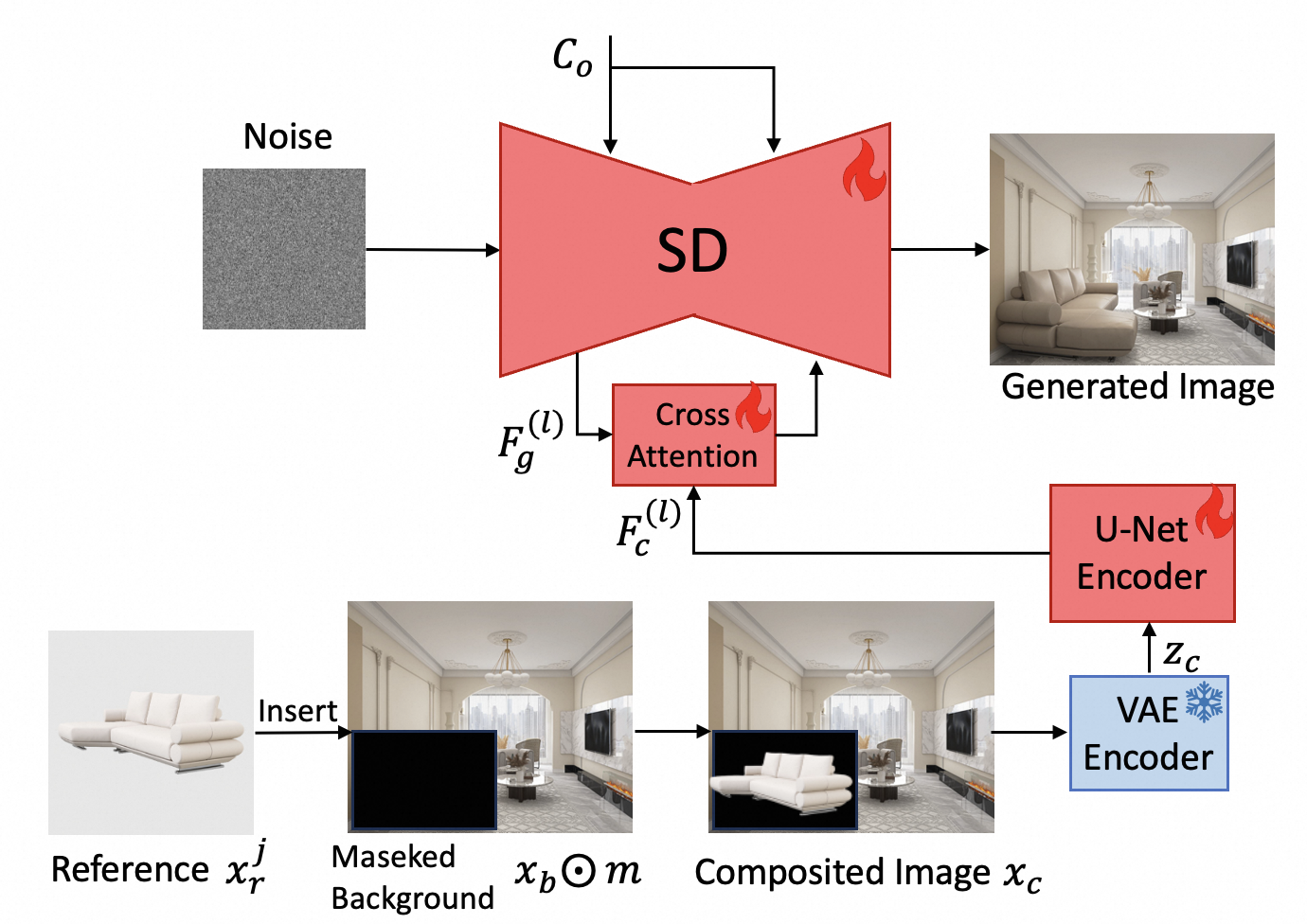}
    \caption{Stage 2: Background-driven Object Customization Learning.}
    \label{fig:1-b}
  \end{subfigure}
  \caption{The overall pipeline of HomeDiffusion. In Stage 1 (MORL), the multi-view object representation $C_o$ is learned from a set of multi-view reference images via an HD visual encoder and an MLP layer. Additionally, a ControlNet is utilized for viewpoint guidance. Stage 2 (BOCL) focuses on integrating the learned multi-view object representation into a background scene at a user-specified location, utilizing a masked background and a composited image to guide the diffusion model in object placement and object details. \textbf{Flames} and \textbf{snowflakes} refer to learnable and frozen parameters, respectively.}
  \label{fig:overview}
\end{figure*}

\section{Methodology}
\subsection{Preliminaries}
\label{sec:prelimi}
Our proposed method is based on the state-of-the-art Stable Diffusion (SD) model, a latent diffusion model that synthesizes high-quality images by modeling the progressive denoising diffusion process in a lower-dimensional latent space. 
Specifically, for an input RGB image $x_0 \in \mathbb{R}^{H\times W\times 3}$, SD first utilize a variational autoencoder (VAE) $\mathcal{E}$ to compress $x_0$ into a smaller low-dimensional latent representation $z_0 \in \mathbb{R}^{h\times w\times c}$. 
Then, during the forward diffusion process, Gaussian noise $\epsilon \sim \mathcal{N}(0,1) $ is progressively added to $z_0$ to obtain the latent representation $z_t$ at time step $t$. 
After that, during the reverse denoising process, a conditional U-Net parameterized by $\theta$ predicts the noise $\epsilon_\theta$ added in the forward process, step by step subtracting the noise $\epsilon_\theta$ until a final latent representation $\hat{z_0}$ is obtained.
Finally, the VAE decoder $\mathcal{D}$ maps $\hat{z_0}$ back to the pixel space, thereby obtaining the generated image $\hat{x_0}$.
During the denoising process, the U-Net can be guided on various forms of condition $C$ such as text prompts or image embeddings through the cross-attention mechanism.
The overall training objective of the SD model can be mathematically expressed as:
\begin{equation}
    \mathcal{L}_{SD} = \mathbb{E}_{z_0, C, \epsilon \sim \mathcal{N}(0,I), t} \left[ \left\| \epsilon - \epsilon_\theta(z_t, t, C) \right\|_{2}^2 \right]
\end{equation}

\begin{figure}[t]
    \centering
    \includegraphics[width=1\columnwidth]{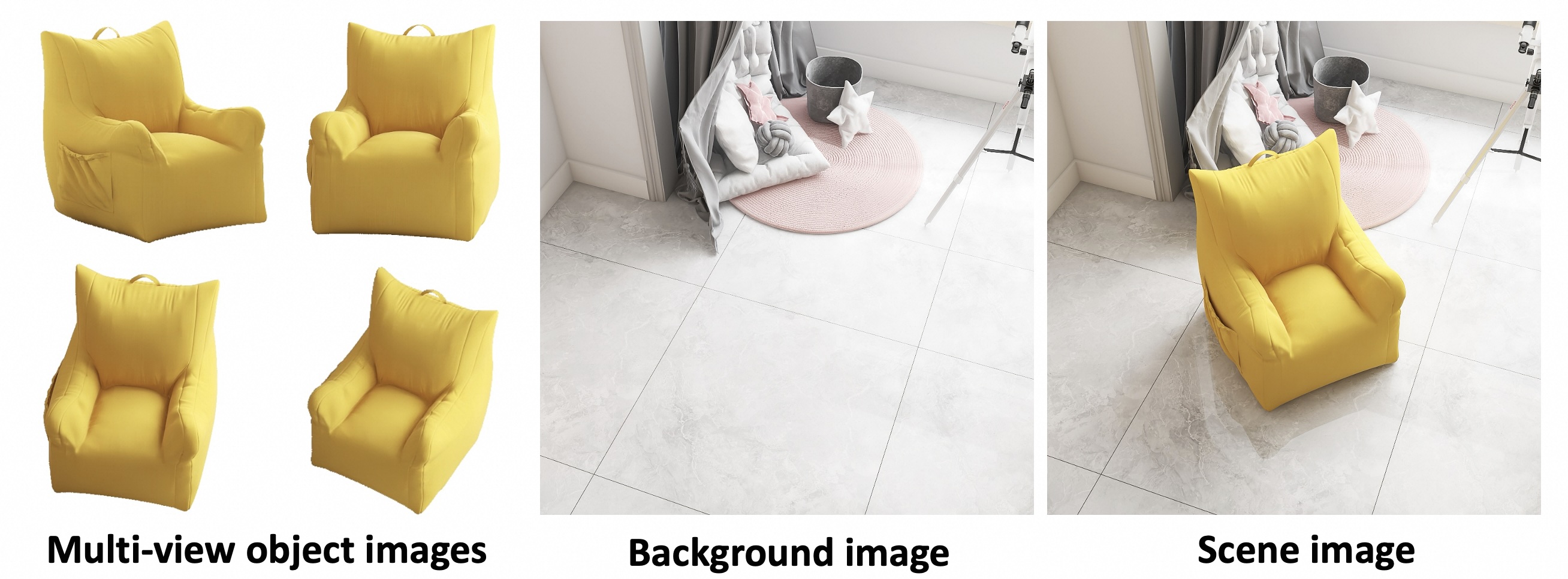}
    \caption{A sample of the training data. }
    \label{fig:datainfo}
\end{figure}

\subsection{Dataset Collection}
\label{sec:data}
A major challenge in object customization image editing is ensuring that the generated object's pose harmonizes with the scene. 
This requires the model to understand the 3D spatial information inherent in 2D images and to adjust the object's perspective relationships appropriately during the generation process. 
Compared to outdoor scenes, indoor scenes have a more defined spatial structure; the presence of walls directly conveys depth and viewpoint information in the images, and the placement and orientation of furniture generally correspond to the layout of the indoor space. Therefore, the intrinsic spatial structure and complexity of indoor scenes pose significant challenges for object customization generative models.
High-quality 3D indoor scene data is scarce, which poses challenges for research. After extensive searching, we found that the 3D-FRONT \cite{fu20203dfront} dataset, commonly used for 3D indoor scene design tasks \cite{hu2024mixeddiffusion3dindoor,yang2024llplace3dindoorscene}, is a suitable dataset.

3D-FRONT is a large-scale, and comprehensive repository of synthetic indoor scenes designed by professional designers. 
We utilized the rendering tool provided by 3D-FRONT to render scene images from different camera angles, where the corresponding furniture will also have different viewpoints. We will segment the target furniture from the images, remove the background, and obtain images of the same furniture from different viewpoints. Then, we can remove the furniture and render a pure background image with the same camera angle, restoring the changes in lighting and shadow caused by placing the furniture. A sample of the constructed training data is shown in Figure \ref{fig:datainfo}.

Specifically, our training dataset contains more than 180k scene images in total, with targeted furniture including 8 categories: bed, chair, coffee table, floor lamp, nightstand, sofa, TV cabinet, and wardrobe. Statistics show that over 60\% of the furniture has images with more than 5 different viewpoints.

\subsection{HomeDiffusion}
\label{sec:overview}
Given $N$ multi-view images $\{x_{r}^{i}\}_{i=1}^{N} \in \mathbb{R}^{h\times w\times 3}$ of a reference object, a background image $x_b \in \mathbb{R}^{H\times W\times 3}$, and a user-specified location binary mask $m \in \mathbb{R}^{H\times W}$ (where 0 pixels represent the background and 1 pixels denote the editable area).
Object customization image editing aims to inject the reference object into the specified area of the background image, ensuring it harmonizes with the background scene.
Facing this task, there are two key points: firstly, the model must be capable of generating any desired viewpoint of the reference object (especially when the object is asymmetric) from a limited set of N reference images $\{x_{r}^{i}\}_{i=1}^{N}$; secondly, the model should generate the reference object with an appropriate pose in the specified area, in accordance with the background layout, ensuring that the generated image is harmonious. 
The overall pipeline of HomeDiffusion is shown in Figure \ref{fig:overview}.
HomeDiffusion consists of two stages: Multi-view Object Representation Learning (MORL) and Background-driven Object Customization Learning (BOCL), which correspond to the two key points mentioned above respectively.

\begin{figure}[t]
    \centering
    \includegraphics[width=1\columnwidth]{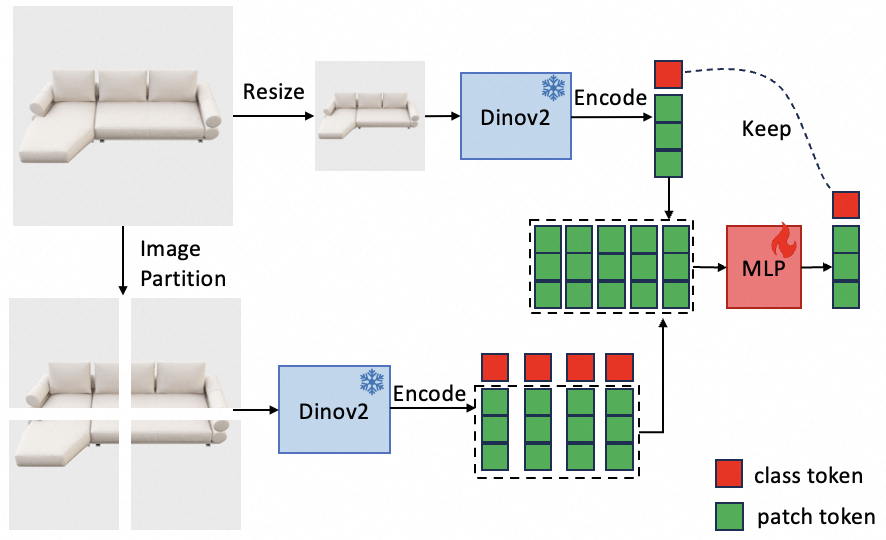}
    \caption{The illustration of HD Visual Encoder. }
    \label{fig:HD}
\end{figure}

\paragraph{HD Visual Encoder.} 
To preserve the details of the object, we need a powerful visual encoder to extract more discriminative features. 
According to the AnyDoor \cite{anydoor} method, we can employ pre-trained DINO-V2 \cite{dinov2} model as the base visual encoder, which encodes a $224\times224$ resolution image as a class token $T_{c}^{1\times1536}$ and patch tokens $T_{p}^{256\times1536}$.
Naturally, we need to be able to encode higher resolution object images, but the number of tokens will sharply increase, leading to a huge demand for GPU memory.
Drawing inspiration from some VLM methods \cite{dong2024internlmxcomposer24khd}, we propose an HD visual encoder that extracts fine visual details from global and local views.
As shown in Figure 4, given a $448\times448$ object image, we process it from both global and local views. 
For the global view, we resize the image to $224\times224$ size and then extract features using DINO-V2. This provides a macro understanding of the image. 
For the local view, we divide the image into 4 non-overlapping $224\times224$ patches, and then extract features for each patch, providing high-resolution local details of the image.
We then need to merge the global and local features.
We concatenate the patch tokens $T_{p}^{5\times256\times1536}$ from all 5 images and then map them to new patch tokens $\hat{T}_{p}^{256\times1536}$ through an MLP layer.
Finally, we retain the class token $T_{c}^{1\times1536}$ from the global view, which provides macro information of object, and together with $\hat{T}_{p}^{256\times1536}$, forms the final high-definition visual encoding. This significantly reduces the number of tokens.

\paragraph{Multi-view Object Representation Learning.} 
Figure \ref{fig:1-a} illustrates the process of MORL.
Given $N$ multi-view reference images $\{x_{r}^{i}\}_{i=1}^{N}$, we randomly select one viewpoint image $x_r^j$ as the source image for the diffusion process, and then use the remaining images $\{x_r^i\}_{i\neq j}$ as the condition to guide the diffusion process.
Each reference image $\{x_r^i\}_{i\neq j}$ will be encoded by the HD visual encoder, yielding multiple object features of size $257 \times 1536$.
We stack these object features and then map them to a $257 \times 1024$ token $C_o$ through an MLP layer. This token, which encapsulates information from multiple reference viewpoints, is subsequently injected into the diffusion process via the cross-attention layers within the U-Net.
But lacking any supplementary information about the target viewpoint, how does the diffusion model generate the target image $x_r^j$ ? 
We employ the ControlNet \cite{controlnet} method to provide extra layout information to guide the image generation process.
Specifically, we extract the Canny edge map $e$ of the target viewpoint $x_r^j$. Then, we use a pre-trained ControlNet model to inject the edge map into the decoder of the U-Net via residual connections, using the layout as a conditional guide for the image generation process. 
Therefore, the training objective of MORL can be expressed as follows:
\begin{equation}
    \mathcal{L}_{MORL} = \mathbb{E}_{z_0, C_o, e, \epsilon} \left[ \left\| \epsilon - \epsilon_\theta(z_t, t, C_o, e) \right\|_{2}^2 \right]
\end{equation}

\paragraph{Background-driven Object Customization Learning.}  
Figure \ref{fig:1-b} shows the BOCL process.
In this stage, we need to train the model to insert the reference object into the specified area of the background image, ensuring that the object's perspective is harmonious with the background scene.
After the MORL stage, we freeze the weights of the MLP layer to preserve the multi-view object representation capabilities of token $C_o$, while also reducing the difficulty of training. 
We combine the background image $x_b$ with the user-specified location mask $m$ to create a masked background image $x_b \odot m$, which provides the scene layout and location information necessary to guide the diffusion model in generating the object with an appropriate viewpoint.
We further paste the reference object image $x_r^j$ onto the editable area in masked background image to create a composited image $x_c$, which contains both scene information and high-fidelity details of the reference object.
We then use the VAE encoder $\mathcal{E}$ to compress the composite image $x_c$ into a latent embedding $z_c$, thus ensuring that the composite image and the generated image are pixel-aligned within the latent space.

Similarly to ControlNet, we utilized a new U-Net encoder to extract multi-level feature maps $F_{c}^{(l)}$ from the latent embedding $z_c$, where $l \in \{1, \ldots, 13\}$. 
Correspondingly, in the SD model, the U-Net encoder will generate multi-level feature maps $F_{g}^{(l)}$ with varying resolutions. 
As Equation \ref{equ:3} shows, we perform cross-attention calculations between $F_{c}^{{(l)}}$ and $F_{g}^{(l)}$ at corresponding levels:
\begin{equation}
\label{equ:3}
\begin{split}
&\text{CrossAttention}(F_{g}^{{(l)}}, F_{c}^{{(l)}}) =\\
& \text{softmax}\left(\frac{(W_Q^{(l)}F_{g}^{{(l)}} (W_K^{(l)}F_{c}^{{(l)}})^T}{\sqrt{d}}\right) W_V^{(l)}F_{c}^{{(l)}},
\end{split}
\end{equation}

where $W_Q^{(l)}$, $W_K^{(l)}$, and $W_V^{(l)}$ represent three learnable projection matrices, and $d$ is the output dimension of the key and query vectors. 
The calculated results will be passed to the corresponding layer of the U-Net decoder.
In this manner, the model can integrate high-fidelity details from the composite image into the generated image at various layer levels.
The training objective of BOCL stage can be expressed as Equation \ref{equ:4}:
\begin{equation}
\label{equ:4}
    \mathcal{L}_{BOCL} = \mathbb{E}_{z_0, x_c, \epsilon}\left[ \left\| \epsilon - \epsilon_\theta(z_t, t, C_o, x_c) \right\|_{2}^2 \right]
\end{equation}

\subsection{Training Strategies}
\label{sec:train}
In the MORL stage, for the multi-view reference image set $\{x_r^i\}_{i\neq j}$, we adopted sampling with replacement, which means that the sampled image set may contain duplicate views. 
Additionally, with a 10\% probability, we randomly select one view image from the reference image set $\{x_r^i\}_{i\neq j}$ and replace all other view images in the set with that one. 
Through this approach, the model is able to learn a robust multi-view object representation. 
Another advantage is that at the inference stage, users can input one or several object reference images, making the model more flexible and convenient to use.
To facilitate classifier-free guidance sampling \cite{cfg}, with a 10\% probability, we set the entire reference image set $\{x_r^i\}_{i\neq j}$ to zero pixel images. 
In the BOCL stage, to drive the model focus on restoring high-fidelity details of the object within the editable area, there is a 50\% probability that we will discard the background area when calculating the loss, called Background Drop. 

\begin{figure}[t]
    \centering
    \includegraphics[width=1\columnwidth]{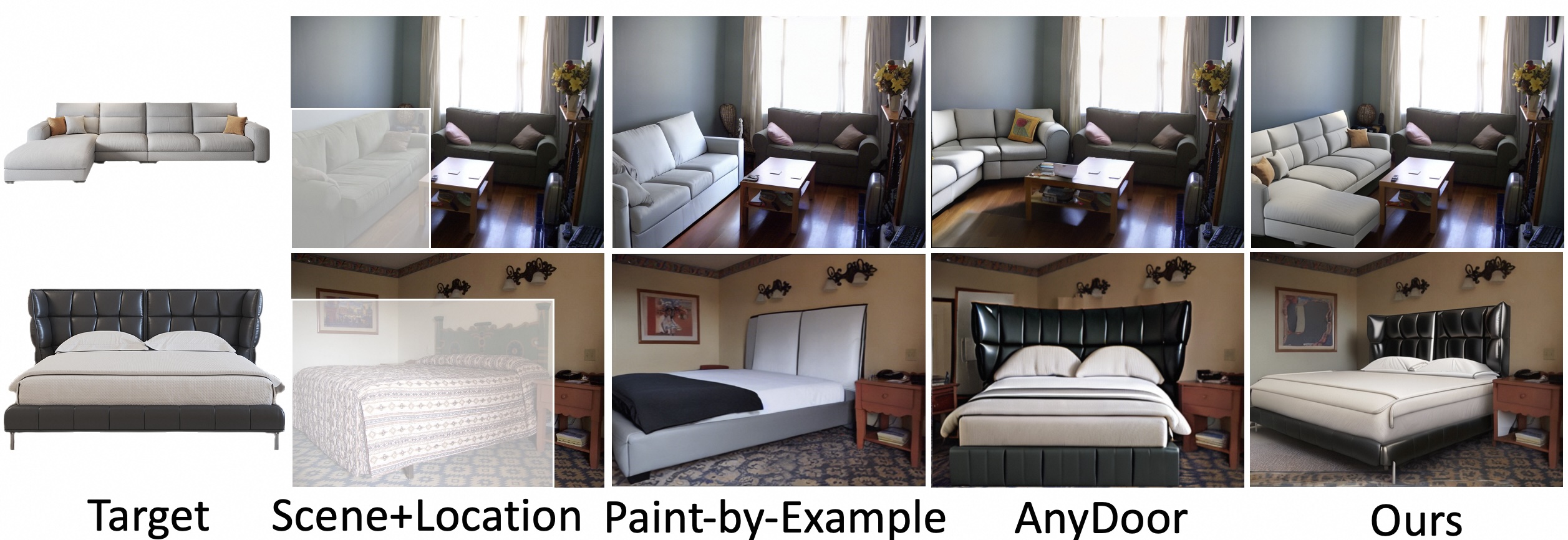}
    \caption{Qualitative comparison with other zero-shot object customization methods. including Paint-by-Example \shortcite{paintbyexample} and AnyDoor \shortcite{anydoor}. \textbf{All methods use only a single-view image with the same resolution as a reference}.}
    \label{fig:4}
\end{figure}

\begin{table}[t]
    \centering
    \begin{tabular}{ccc}
    \toprule
    Method & CLIP Score  & DINO Score \\
    \midrule
    Paint-by-Example \shortcite{paintbyexample} & 83.4 & 78.5 \\
    AnyDoor \shortcite{anydoor} & 87.1 & 83.3 \\
    Ours & \textbf{89.4} & \textbf{86.2} \\
    \bottomrule
    \end{tabular}
    \caption{Quantitative comparison of different zero-shot methods on ZOC-Indoor-Eval benchmark using single viewpoint reference image.}
    \label{tab:1}
\end{table}

\section{Experiments}
\subsection{Implementation Details and Evaluation}
\paragraph{Implementation Details.}
We use Stable Diffusion V2.1 as our backbone, and choose DINO-V2 giant version as the base image encoder. 
The resolution of the training images is $512 \times 512$.
We use the Adam optimizer \shortcite{Adam} with the initial learning rate of $1e^{-5}$ and a weight decay of $1e^{-2}$ to train our model. 
We train our model on 8 NVIDIA Tesla V100 GPUs using PyTorch \cite{pytorch}. 
\paragraph{Benchmarks.}
For quantitative results, we construct a new benchmark, named the ZOC-Indoor-Val benchmark.
It comprises 8 categories of furniture (described in the Dataset Collection Section).
Each category includes 20 different scenes, where a scene consists of a multi-view reference image set and a scene file.
The multi-view reference image set contains 5 images of the furniture from different views, while the scene file includes a scene image and a binary mask indicating the furniture's location.
The view of the furniture in the scene image is different from all reference images, providing a means to validate whether the model can understand the spatial and perspective relationships between the object and its surroundings.
We plan to publicly release the ZOC-Indoor-Val benchmark.
To facilitate comparison with other relevant methods, we also performed quantitative analysis on the Viton-HD test \cite{viton} benchmark to measure the performance of different models in virtual try-on.
\textbf{In all qualitative and quantitative experiments, the resolution of the reference images is the same.}
\paragraph{Evaluation Metrics.}
Following DreamBooth \cite{ruiz2023dreambooth} and AnyDoor \cite{anydoor} methods, we use the CLIP Score and DINO Score to calculate the similarity between the generated object and the actual target object in the scene image, where higher scores indicate better performance. 

\begin{table}[t]
    \centering
    \begin{tabular}{ccc}
    \toprule
     Method & CLIP Score  & DINO Score  \\
    \midrule
    Paint-by-Example \shortcite{paintbyexample} & 80.1 & 56.7 \\
    AnyDoor \shortcite{anydoor} & 81.8 & 59.3 \\
    Ours  & \textbf{82.2} & \textbf{60.3} \\
    \bottomrule
    \end{tabular}
    \caption{Quantitative comparison of different zero-shot methods on the Viton-HD test benchmark using single viewpoint reference image.}
    \label{tab:3}
\end{table}

\subsection{Comparisons with Related Methods}

\paragraph{Comparison with Methods Using a Single-view Object Image.}
The results from Table \ref{tab:1} clearly demonstrate the superior performance of our method in comparison to existing zero-shot methods, as evaluated on the ZOC-Indoor-Eval benchmark. Our approach outperforms the AnyDoor and Paint-by-Example methods, achieving a CLIP Score of 89.4 and a DINO Score of 86.2.
In Figure \ref{fig:4}, we present the visualization results compared with Paint-by-Example and AnyDoor.
It is evident that while Paint-by-Example can generate the correct perspective for the target bed, it fails to preserve details. 
Because Paint-by-Example employs CLIP to encode reference images and utilizes only the class token portion as the condition.
As stated in \cite{paintbyexample}: "This tends to ignore the high-frequency details".
AnyDoor verified that using DINO-V2 as the image encoder and taking all tokens as the condition significantly improves fidelity.
However, due to the lack of multi-view object representation information, AnyDoor appears to simply "paste" the target bed into the image, resulting in a discordant perspective. Our method, on the other hand, not only generates the appropriate perspective but also retains the details well. 
Although we add an encoding network, it requires only a single-step computation, with the extracted multi-level features being utilized in each subsequent diffusion step. Specifically, for the 20-step diffusion sampling at a 512 resolution, it adds only an extra 5\% to the inference time.

\begin{table}[t]
    \centering
    \setlength{\tabcolsep}{1mm}
    \begin{tabular}{cccc}
    \toprule
    Method & One view & Three views & Five views\\
    \midrule
    DreamBooth \shortcite{ruiz2023dreambooth} & 69.8 & 81.3 & 85.5\\
    Ours  & 86.2 & 87.5 & 88.1 \\
    \bottomrule
    \end{tabular}
    \caption{Quantitative comparison of our method with the DreamBooth method on the ZOC-Indoor-Eval benchmark. The terms 'one,' 'three,' and 'five' views correspond to using one, three, and five different viewpoint reference images, respectively. The reported results are the DINO Scores.}
    \label{tab:4}
\end{table}

\begin{figure}[t]
    \centering
    \includegraphics[width=0.9\columnwidth]{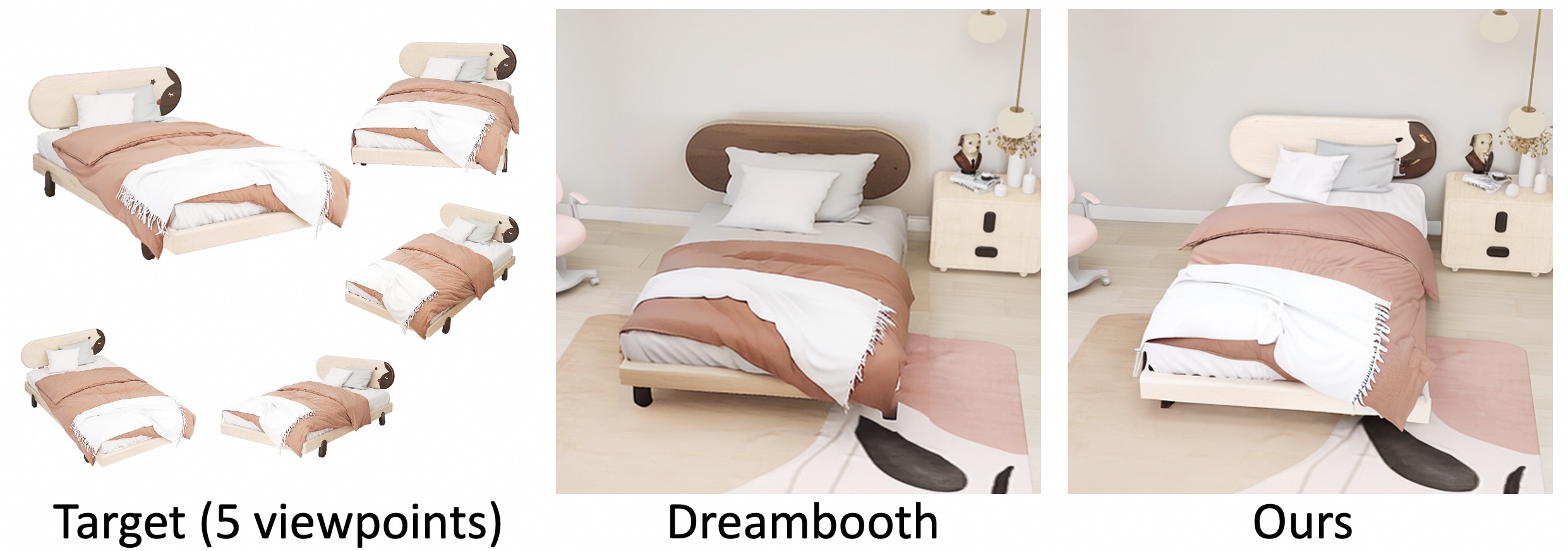}
    \caption{Qualitative comparison with DreamBooth. Our method achieves higher fidelity in details such as the color and pattern of the headboard.}
    \label{fig:HomeDiffusion-dreambooth}
\end{figure}

\begin{table}[t]
    \centering
    \begin{tabular}{ccc}
    \toprule
    Method & Fidelity ($\uparrow$) & Harmony ($\uparrow$) \\
    \midrule
    DreamBooth \shortcite{ruiz2023dreambooth} & 2.18 & 2.57 \\
    AnyDoor \shortcite{anydoor} & 2.70 & 2.46 \\
    Ours & 3.45 & 3.51 \\
    \bottomrule
    \end{tabular}
    \caption{The average results of the small-scale human evaluation study. "Fidelity" measures object identity preservation, and "Harmony" evaluates the object's consistency with its surroundings in terms of viewpoint and lighting.}
    \label{tab:user-study}
\end{table}

\paragraph{Comparison with Methods Using Multi-view Object Images.}
To validate the effectiveness of our method in leveraging multi-viewpoint object information, we compare our method with the few-shot DreamBooth \cite{ruiz2023dreambooth} approach.
DreamBooth accepts one or more images of the same object and then undergoes extensive training to link the object concept to a specific word for generation during inference.
Our method, conversely, requires no such training and can efficiently utilize unseen images from multiple viewpoints to generate the desired object.
Table \ref{tab:4} presents the quantitative results. 
Our method outperforms the DreamBooth method under the conditions of "one viewpoint," "three viewpoints," and "five viewpoints," achieving DINO Scores of 86.2, 87.5, and 88.1, respectively. As the number of received viewpoints increases, the DINO Score also progressively improves, indicating the effectiveness of our method in utilizing multi-view object information.
Figure \ref{fig:HomeDiffusion-dreambooth} shows our method's fidelity advantage over DreamBooth.

\paragraph{User Study.}
We invited 6 participants, including interior design experts and novices, to score images generated by different methods.
A total of 900 images were evaluated on "Fidelity" and "Harmony".
"Fidelity" measures the maintenance of object identity, and "Harmony" assesses congruence with surrounding viewpoints and lighting.
Scores ranged from 1 (lowest) to 5 (highest), with the average score for each method calculated. As shown in Table \ref{tab:user-study}, our method demonstrated significant advantages in both fidelity and harmony compared to AnyDoor and DreamBooth.

\begin{figure}[t]
    \centering
    \includegraphics[width=1\columnwidth]{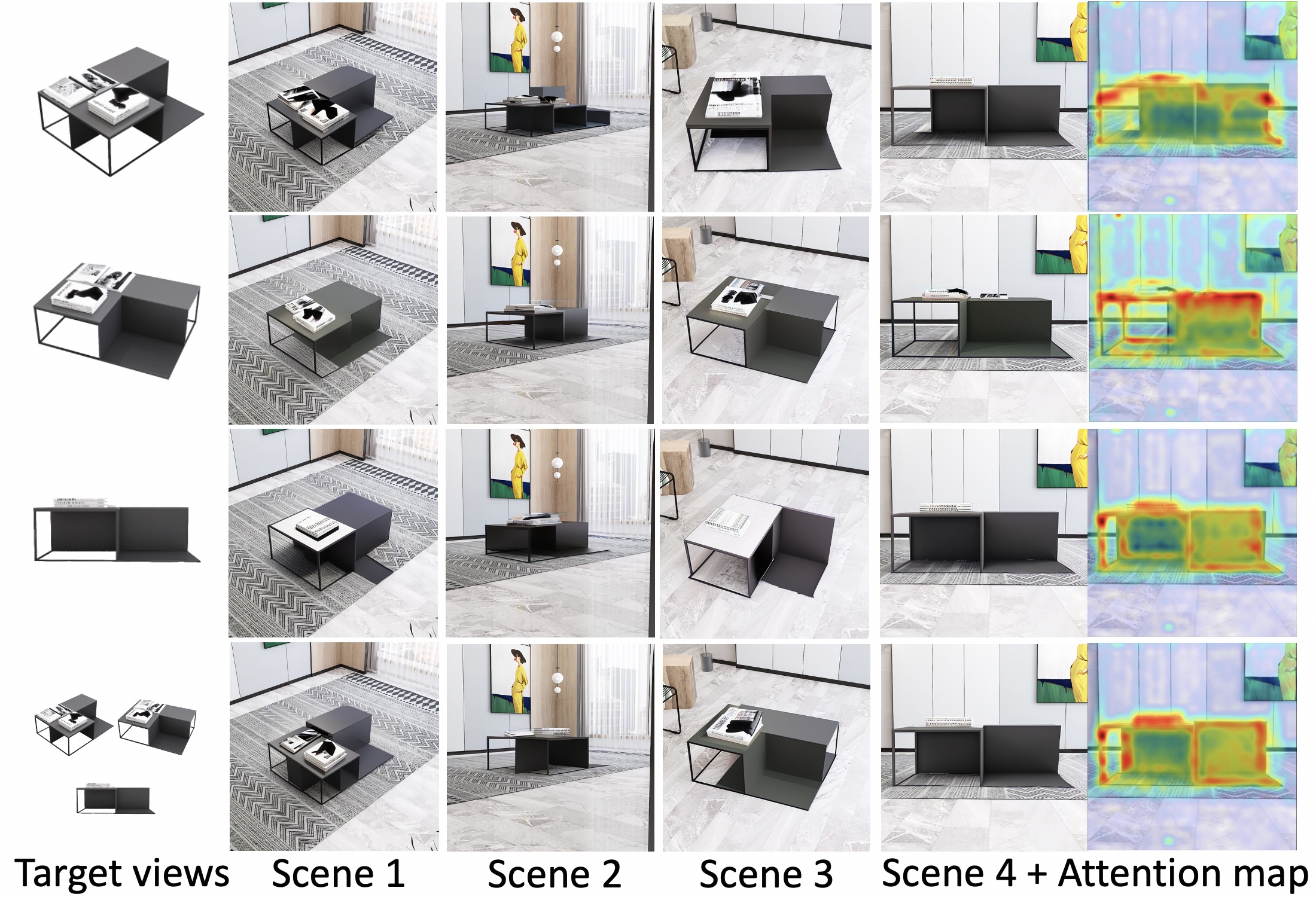}
    \caption{The impact of target viewpoints ablation on HomeDiffusion performance. The first to third rows show the generation results under different single viewpoint of the furniture, while the fourth row depicts the results when all three viewpoints are provided. In Scene 4, we specifically visualize the attention map within the U-Net.}
    \label{fig:multiview-compare}
\end{figure}

\begin{table}[t]
    \centering
    \begin{tabular}{ccc}
    \toprule
      & One view & Five views \\
    \midrule
    w/o MORL & 84.6 & 86.9\\
    Full  & 86.2  & 88.1 \\
    \bottomrule
    \end{tabular}
    \caption{Ablation studies on the MORL method. The results shown in the table are the DINO Scores measured on the ZOC-Indoor-Eval benchmark.}
    \label{tab:5}
\end{table}

\subsection{Ablation Study}
\paragraph{Analysis on Multi-view Object Representation Learning.}
To verify the effectiveness of our proposed MORL method, we conduct ablation experiments reported in Figure \ref{fig:multiview-compare} and Table \ref{tab:5}.
In Figure \ref{fig:multiview-compare}, we conducted an ablation test on a complex structured piece of furniture, which cannot be accurately reconstructed from a single viewpoint alone. We then chose four scenes with different orientations to observe the generation results when only a single viewpoint is provided, finally comparing these with the results when all three viewpoint images are available. It can be observed in detail that when there is a significant difference between the furniture's viewpoint and the scene's orientation, reconstruction becomes challenging, leading to various issues such as color changes and panels turning into hollow spaces. However, when HomeDiffusion receives all three viewpoints, it can effectively understand the furniture's multi-view information and generate accurate results in different scenes.
We further visualized the attention map within the U-Net for the results of Scene 4.
It is noticeable that the attention map in the third row is most similar to the one when multiple viewpoints are used, and the furniture view in the third row closely matches the orientation of Scene 4.
This demonstrates HomeDiffusion's capability in correlating the viewpoint of the objects with the scene.
The quantitative results in Table \ref{tab:5} further suggest that MORL learns effective multi-view object information through a self-generative training process. The absence of this initial information is likely to increase the difficulty of subsequent BOCL training.
\begin{figure}[t]
    \centering
    \includegraphics[width=1\columnwidth]{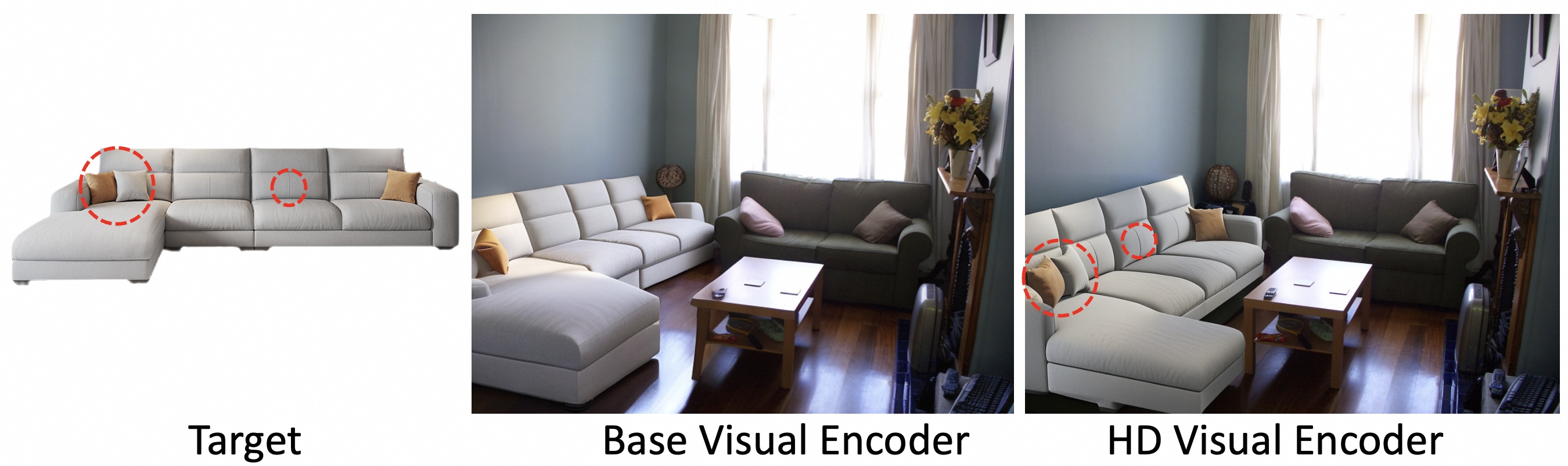}
    \caption{Ablation on the impact of HD Visual Encoder.}
    \label{fig:exp-hd}
\end{figure}

\begin{table}[t]
    \centering
    \setlength{\tabcolsep}{1mm}
    \begin{tabular}{lcc}
    \toprule
     & One view  & Five views\\
    \midrule
    Baseline & 83.8 & 85.0\\
    $+$ HD visual encoder  & 85.2  & 86.3 \\
    $+$ Composited image guidance  & 85.6  & 86.9 \\
    $+$ Pixel-aligned cross-attention  & 86.2  & 88.1 \\
  \bottomrule
    \end{tabular}
    \caption{Ablation studies on the high-fidelity object details extraction. The results shown in the table are the DINO Scores measured on the ZOC-Indoor-Eval benchmark.}
    \label{tab:6}
\end{table}

\paragraph{Analysis on High-fidelity Object Details Extraction.}
Table \ref{tab:6} presents the ablation studies on the high-fidelity detail extraction. 
The baseline contains the complete MORL stage, but for the BOCL process, we used the masked image $x_b \odot m$ alone as the guiding condition.
Compared to the baseline, we introduced the HD visual encoder to replace the basic DINO-V2 encoder, which resulted in significant score increases of 1.4 and 1.3 in one and five views, respectively.
This gain can be visually seen in Figure \ref{fig:exp-hd}, where the HD visual encoder can extract more fine details, such as the correct number of pillows and the linings on the sofa.
The composited image guidance strategy means that we paste the reference object  $x_r^j$ into the masked image and then use this composite image as the guiding condition, thereby transferring the rich high-fidelity details from the reference image into the diffusion model.
With this strategy, the DINO scores for one view case and five views case increased by 0.4 and 0.6 points, respectively.
When the composite image is encoded into the latent space, because the pasted reference object’s viewpoint does not align with the target viewpoint in the generated image, directly adding the feature maps from each level could have a negative effect on the results. 
Instead, applying pixel-aligned cross-attention computations for each level of feature maps will be more effective. This method brings 0.6 and 1.2 DINO score gains for cases with one view and five views, respectively.

\begin{figure}[t]
    \centering
    \includegraphics[width=0.85\columnwidth]{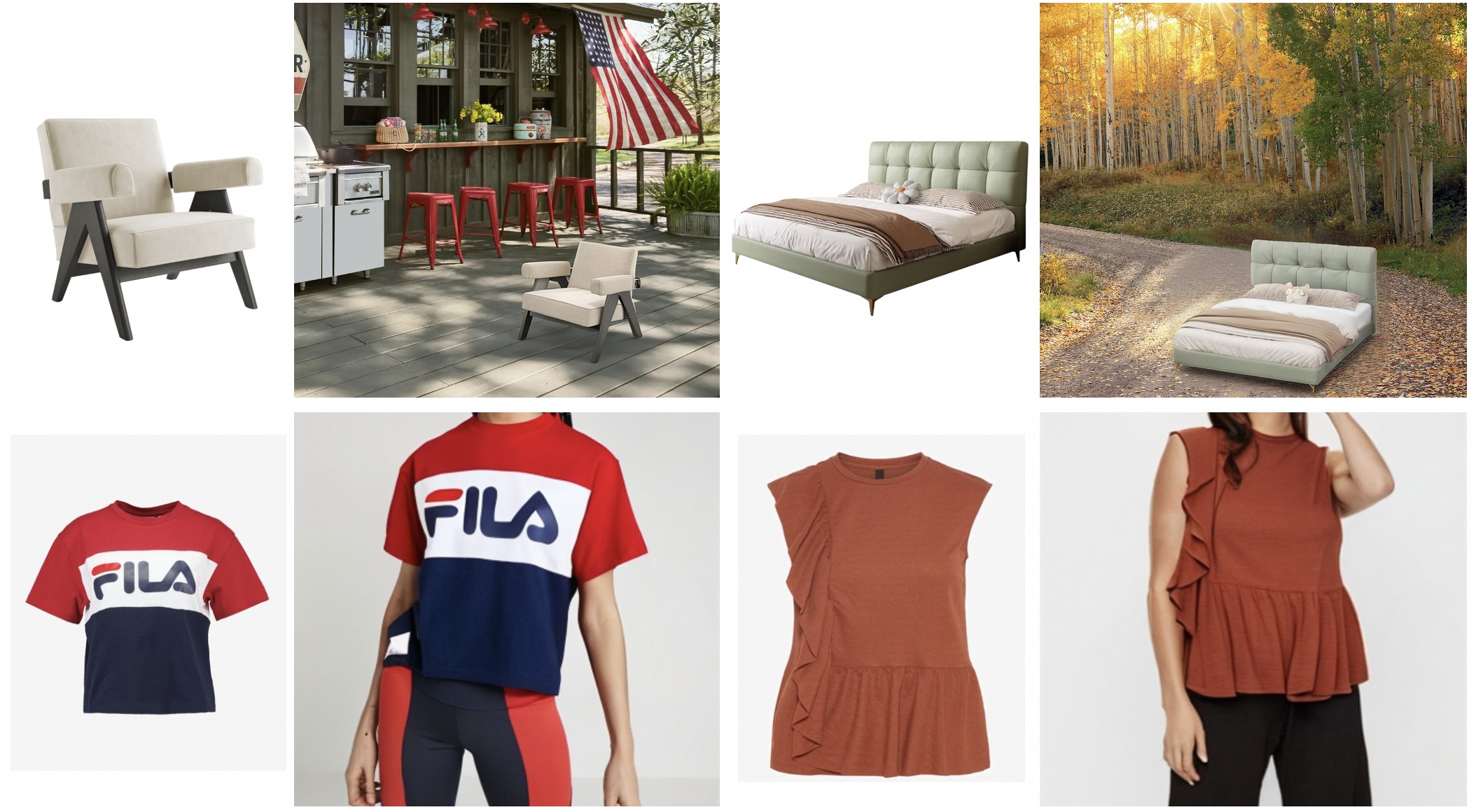}
    \caption{The generalization ability of HomeDiffusion.}
    \label{fig:general-all}
\end{figure}

\subsection{The Generalization Ability of HomeDiffusion}
Although the training data for HomeDiffusion all come from indoor spaces, it can also generalize to be used in outdoor natural scenes. As shown in the first row of Figure \ref{fig:general-all}, chairs and beds can be naturally placed in outdoor scenes with high fidelity and harmony.
We also trained our network on the Viton-HD dataset. The results, as shown in the second row of Figure \ref{fig:general-all}, demonstrate that our method performs well in virtual try-on scenarios. Furthermore, when comparing our method to others using the Viton-HD-test dataset, as Table \ref{tab:3} shows, it is evident that our method outperforms AnyDoor and Paint-by-Example. This underscores the potential of our method to be applied across a broader range of fields.

\section{Conclusion}
In this work, we present HomeDiffusion, a novel zero-shot object customization method that maintains detailed fidelity of objects while seamlessly blending them into background scenes. We propose an HD visual encoder to extract fine image details from global and local views. We also introduce a MORL method, which learns multi-view object representations through self-generative training from various views. Moreover, we propose a BOCL approach that skillfully incorporates objects into specific areas within scenes. Our experimental results demonstrated the good performance of HomeDiffusion in customizing objects within indoor scenes, as well as its potential to generalize to other scenes.

\bibliography{aaai25}

@misc{rombach2021highresolution,
      title={High-Resolution Image Synthesis with Latent Diffusion Models}, 
      author={Robin Rombach and Andreas Blattmann and Dominik Lorenz and Patrick Esser and Björn Ommer},
      year={2021},
      eprint={2112.10752},
      archivePrefix={arXiv},
      primaryClass={cs.CV}
}

@inproceedings{radford2021learning,
  title={Learning transferable visual models from natural language supervision},
  author={Radford, Alec and Kim, Jong Wook and Hallacy, Chris and Ramesh, Aditya and Goh, Gabriel and Agarwal, Sandhini and Sastry, Girish and Askell, Amanda and Mishkin, Pamela and Clark, Jack and others},
  booktitle={International conference on machine learning},
  pages={8748--8763},
  year={2021},
  organization={PMLR}
}

@inproceedings{controlnet,
  title={Adding conditional control to text-to-image diffusion models},
  author={Zhang, Lvmin and Rao, Anyi and Agrawala, Maneesh},
  booktitle={Proceedings of the IEEE/CVF International Conference on Computer Vision},
  pages={3836--3847},
  year={2023}
}

@misc{dinov2,
      title={DINOv2: Learning Robust Visual Features without Supervision}, 
      author={Maxime Oquab and Timothée Darcet and Théo Moutakanni and Huy Vo and Marc Szafraniec and Vasil Khalidov and Pierre Fernandez and Daniel Haziza and Francisco Massa and Alaaeldin El-Nouby and Mahmoud Assran and Nicolas Ballas and Wojciech Galuba and Russell Howes and Po-Yao Huang and Shang-Wen Li and Ishan Misra and Michael Rabbat and Vasu Sharma and Gabriel Synnaeve and Hu Xu and Hervé Jegou and Julien Mairal and Patrick Labatut and Armand Joulin and Piotr Bojanowski},
      year={2024},
      eprint={2304.07193},
      archivePrefix={arXiv},
      primaryClass={cs.CV},
      url={https://arxiv.org/abs/2304.07193}, 
}

@inproceedings{anydoor,
  title={Anydoor: Zero-shot object-level image customization},
  author={Chen, Xi and Huang, Lianghua and Liu, Yu and Shen, Yujun and Zhao, Deli and Zhao, Hengshuang},
  booktitle={Proceedings of the IEEE/CVF Conference on Computer Vision and Pattern Recognition},
  pages={6593--6602},
  year={2024}
}

@misc{cfg,
      title={Classifier-Free Diffusion Guidance}, 
      author={Jonathan Ho and Tim Salimans},
      year={2022},
      eprint={2207.12598},
      archivePrefix={arXiv},
      primaryClass={cs.LG},
      url={https://arxiv.org/abs/2207.12598}, 
}

@misc{2021GLIDE,
      title={GLIDE: Towards Photorealistic Image Generation and Editing with Text-Guided Diffusion Models}, 
      author={Alex Nichol and Prafulla Dhariwal and Aditya Ramesh and Pranav Shyam and Pamela Mishkin and Bob McGrew and Ilya Sutskever and Mark Chen},
      year={2022},
      eprint={2112.10741},
      archivePrefix={arXiv},
      primaryClass={cs.CV},
      url={https://arxiv.org/abs/2112.10741}, 
}

@misc{Dai2023EmuEI,
      title={Emu: Enhancing Image Generation Models Using Photogenic Needles in a Haystack}, 
      author={Xiaoliang Dai and Ji Hou and Chih-Yao Ma and Sam Tsai and Jialiang Wang and Rui Wang and Peizhao Zhang and Simon Vandenhende and Xiaofang Wang and Abhimanyu Dubey and Matthew Yu and Abhishek Kadian and Filip Radenovic and Dhruv Mahajan and Kunpeng Li and Yue Zhao and Vladan Petrovic and Mitesh Kumar Singh and Simran Motwani and Yi Wen and Yiwen Song and Roshan Sumbaly and Vignesh Ramanathan and Zijian He and Peter Vajda and Devi Parikh},
      year={2023},
      eprint={2309.15807},
      archivePrefix={arXiv},
      primaryClass={cs.CV},
      url={https://arxiv.org/abs/2309.15807}, 
}

@article{DiffuisonBeatGAN,
  title={Diffusion models beat gans on image synthesis},
  author={Dhariwal, Prafulla and Nichol, Alexander},
  journal={Advances in neural information processing systems},
  volume={34},
  pages={8780--8794},
  year={2021}
}

@article{ho2020denoising,
  title={Denoising diffusion probabilistic models},
  author={Ho, Jonathan and Jain, Ajay and Abbeel, Pieter},
  journal={Advances in neural information processing systems},
  volume={33},
  pages={6840--6851},
  year={2020}
}

@article{Imagen,
  title={Photorealistic text-to-image diffusion models with deep language understanding},
  author={Saharia, Chitwan and Chan, William and Saxena, Saurabh and Li, Lala and Whang, Jay and Denton, Emily L and Ghasemipour, Kamyar and Gontijo Lopes, Raphael and Karagol Ayan, Burcu and Salimans, Tim and others},
  journal={Advances in Neural Information Processing Systems},
  volume={35},
  pages={36479--36494},
  year={2022}
}

@inproceedings{paintbyexample,
  title={Paint by example: Exemplar-based image editing with diffusion models},
  author={Yang, Binxin and Gu, Shuyang and Zhang, Bo and Zhang, Ting and Chen, Xuejin and Sun, Xiaoyan and Chen, Dong and Wen, Fang},
  booktitle={Proceedings of the IEEE/CVF Conference on Computer Vision and Pattern Recognition},
  pages={18381--18391},
  year={2023}
}

@inproceedings{repaint,
  title={Repaint: Inpainting using denoising diffusion probabilistic models},
  author={Lugmayr, Andreas and Danelljan, Martin and Romero, Andres and Yu, Fisher and Timofte, Radu and Van Gool, Luc},
  booktitle={Proceedings of the IEEE/CVF Conference on Computer Vision and Pattern Recognition},
  pages={11461--11471},
  year={2022}
}

@misc{sdedit,
      title={SDEdit: Guided Image Synthesis and Editing with Stochastic Differential Equations}, 
      author={Chenlin Meng and Yutong He and Yang Song and Jiaming Song and Jiajun Wu and Jun-Yan Zhu and Stefano Ermon},
      year={2022},
      eprint={2108.01073},
      archivePrefix={arXiv},
      primaryClass={cs.CV},
      url={https://arxiv.org/abs/2108.01073}, 
}

@inproceedings{ruiz2023dreambooth,
  title={Dreambooth: Fine tuning text-to-image diffusion models for subject-driven generation},
  author={Ruiz, Nataniel and Li, Yuanzhen and Jampani, Varun and Pritch, Yael and Rubinstein, Michael and Aberman, Kfir},
  booktitle={Proceedings of the IEEE/CVF Conference on Computer Vision and Pattern Recognition},
  pages={22500--22510},
  year={2023}
}

@misc{Adam,
      title={Adam: A Method for Stochastic Optimization}, 
      author={Diederik P. Kingma and Jimmy Ba},
      year={2017},
      eprint={1412.6980},
      archivePrefix={arXiv},
      primaryClass={cs.LG},
      url={https://arxiv.org/abs/1412.6980}, 
}

@article{pytorch,
  title={Pytorch: An imperative style, high-performance deep learning library},
  author={Paszke, Adam and Gross, Sam and Massa, Francisco and Lerer, Adam and Bradbury, James and Chanan, Gregory and Killeen, Trevor and Lin, Zeming and Gimelshein, Natalia and Antiga, Luca and others},
  journal={Advances in neural information processing systems},
  volume={32},
  pages={8026--8037},
  year={2019}
}

@misc{fu20203dfront,
      title={3D-FRONT: 3D Furnished Rooms with layOuts and semaNTics}, 
      author={Huan Fu and Bowen Cai and Lin Gao and Lingxiao Zhang and Jiaming Wang Cao Li and Zengqi Xun and Chengyue Sun and Rongfei Jia and Binqiang Zhao and Hao Zhang},
      year={2021},
      eprint={2011.09127},
      archivePrefix={arXiv},
      primaryClass={cs.CV},
      url={https://arxiv.org/abs/2011.09127}, 
}

@inproceedings{viton,
  title={VITON-HD: High-Resolution Virtual Try-On via Misalignment-Aware Normalization},
  author={Choi, Seunghwan and Park, Sunghyun and Lee, Minsoo and Choo, Jaegul},
  booktitle={Proc. of the IEEE conference on computer vision and pattern recognition (CVPR)},
  year={2021}
}

@misc{shi2023instantbooth,
      title={InstantBooth: Personalized Text-to-Image Generation without Test-Time Finetuning}, 
      author={Jing Shi and Wei Xiong and Zhe Lin and Hyun Joon Jung},
      year={2023},
      eprint={2304.03411},
      archivePrefix={arXiv},
      primaryClass={cs.CV},
      url={https://arxiv.org/abs/2304.03411}, 
}

@inproceedings{CustomDiffusion,
  title={Multi-concept customization of text-to-image diffusion},
  author={Kumari, Nupur and Zhang, Bingliang and Zhang, Richard and Shechtman, Eli and Zhu, Jun-Yan},
  booktitle={Proceedings of the IEEE/CVF Conference on Computer Vision and Pattern Recognition},
  pages={1931--1941},
  year={2023}
}

@misc{worthoneword,
      title={An Image is Worth One Word: Personalizing Text-to-Image Generation using Textual Inversion}, 
      author={Rinon Gal and Yuval Alaluf and Yuval Atzmon and Or Patashnik and Amit H. Bermano and Gal Chechik and Daniel Cohen-Or},
      year={2022},
      eprint={2208.01618},
      archivePrefix={arXiv},
      primaryClass={cs.CV},
      url={https://arxiv.org/abs/2208.01618}, 
}

@article{SuTI,
  title={Subject-driven text-to-image generation via apprenticeship learning},
  author={Chen, Wenhu and Hu, Hexiang and Li, Yandong and Ruiz, Nataniel and Jia, Xuhui and Chang, Ming-Wei and Cohen, William W},
  journal={Advances in Neural Information Processing Systems},
  volume={36},
  year={2024}
}

@misc{xiao2023fastcomposer,
      title={FastComposer: Tuning-Free Multi-Subject Image Generation with Localized Attention}, 
      author={Guangxuan Xiao and Tianwei Yin and William T. Freeman and Frédo Durand and Song Han},
      year={2023},
      eprint={2305.10431},
      archivePrefix={arXiv},
      primaryClass={cs.CV},
      url={https://arxiv.org/abs/2305.10431}, 
}

@misc{podell2023sdxl,
      title={SDXL: Improving Latent Diffusion Models for High-Resolution Image Synthesis}, 
      author={Dustin Podell and Zion English and Kyle Lacey and Andreas Blattmann and Tim Dockhorn and Jonas Müller and Joe Penna and Robin Rombach},
      year={2023},
      eprint={2307.01952},
      archivePrefix={arXiv},
      primaryClass={cs.CV},
      url={https://arxiv.org/abs/2307.01952}, 
}

@misc{prompt2prompt,
      title={Prompt-to-Prompt Image Editing with Cross Attention Control}, 
      author={Amir Hertz and Ron Mokady and Jay Tenenbaum and Kfir Aberman and Yael Pritch and Daniel Cohen-Or},
      year={2022},
      eprint={2208.01626},
      archivePrefix={arXiv},
      primaryClass={cs.CV},
      url={https://arxiv.org/abs/2208.01626}, 
}

@inproceedings{kawar2023imagic,
  title={Imagic: Text-based real image editing with diffusion models},
  author={Kawar, Bahjat and Zada, Shiran and Lang, Oran and Tov, Omer and Chang, Huiwen and Dekel, Tali and Mosseri, Inbar and Irani, Michal},
  booktitle={Proceedings of the IEEE/CVF Conference on Computer Vision and Pattern Recognition},
  pages={6007--6017},
  year={2023}
}

@inproceedings{kim2022diffusionclip,
  title={Diffusionclip: Text-guided diffusion models for robust image manipulation},
  author={Kim, Gwanghyun and Kwon, Taesung and Ye, Jong Chul},
  booktitle={Proceedings of the IEEE/CVF Conference on Computer Vision and Pattern Recognition},
  pages={2426--2435},
  year={2022}
}

@misc{balaji2022ediffi,
      title={eDiff-I: Text-to-Image Diffusion Models with an Ensemble of Expert Denoisers}, 
      author={Yogesh Balaji and Seungjun Nah and Xun Huang and Arash Vahdat and Jiaming Song and Qinsheng Zhang and Karsten Kreis and Miika Aittala and Timo Aila and Samuli Laine and Bryan Catanzaro and Tero Karras and Ming-Yu Liu},
      year={2023},
      eprint={2211.01324},
      archivePrefix={arXiv},
      primaryClass={cs.CV},
      url={https://arxiv.org/abs/2211.01324}, 
}

@misc{li2023photomaker,
      title={PhotoMaker: Customizing Realistic Human Photos via Stacked ID Embedding}, 
      author={Zhen Li and Mingdeng Cao and Xintao Wang and Zhongang Qi and Ming-Ming Cheng and Ying Shan},
      year={2023},
      eprint={2312.04461},
      archivePrefix={arXiv},
      primaryClass={cs.CV},
      url={https://arxiv.org/abs/2312.04461}, 
}

@misc{voynov2023p+,
      title={P+: Extended Textual Conditioning in Text-to-Image Generation}, 
      author={Andrey Voynov and Qinghao Chu and Daniel Cohen-Or and Kfir Aberman},
      year={2023},
      eprint={2303.09522},
      archivePrefix={arXiv},
      primaryClass={cs.CV},
      url={https://arxiv.org/abs/2303.09522}, 
}

@misc{chen2022reimage,
      title={Re-Imagen: Retrieval-Augmented Text-to-Image Generator}, 
      author={Wenhu Chen and Hexiang Hu and Chitwan Saharia and William W. Cohen},
      year={2022},
      eprint={2209.14491},
      archivePrefix={arXiv},
      primaryClass={cs.CV},
      url={https://arxiv.org/abs/2209.14491}, 
}

@article{goodfellow2020generative,
  title={Generative adversarial networks},
  author={Goodfellow, Ian and Pouget-Abadie, Jean and Mirza, Mehdi and Xu, Bing and Warde-Farley, David and Ozair, Sherjil and Courville, Aaron and Bengio, Yoshua},
  journal={Communications of the ACM},
  volume={63},
  number={11},
  pages={139--144},
  year={2020},
  publisher={ACM New York, NY, USA}
}

@article{ho2022cascaded,
  title={Cascaded diffusion models for high fidelity image generation},
  author={Ho, Jonathan and Saharia, Chitwan and Chan, William and Fleet, David J and Norouzi, Mohammad and Salimans, Tim},
  journal={The Journal of Machine Learning Research},
  volume={23},
  number={1},
  pages={2249--2281},
  year={2022},
  publisher={JMLRORG}
}

@inproceedings{kumari2023ablating,
  title={Ablating concepts in text-to-image diffusion models},
  author={Kumari, Nupur and Zhang, Bingliang and Wang, Sheng-Yu and Shechtman, Eli and Zhang, Richard and Zhu, Jun-Yan},
  booktitle={Proceedings of the IEEE/CVF International Conference on Computer Vision},
  pages={22691--22702},
  year={2023}
}

@inproceedings{zhang2023sine,
  title={Sine: Single image editing with text-to-image diffusion models},
  author={Zhang, Zhixing and Han, Ligong and Ghosh, Arnab and Metaxas, Dimitris N and Ren, Jian},
  booktitle={Proceedings of the IEEE/CVF Conference on Computer Vision and Pattern Recognition},
  pages={6027--6037},
  year={2023}
}

@misc{hu2024mixeddiffusion3dindoor,
      title={Mixed Diffusion for 3D Indoor Scene Synthesis}, 
      author={Siyi Hu and Diego Martin Arroyo and Stephanie Debats and Fabian Manhardt and Luca Carlone and Federico Tombari},
      year={2024},
      eprint={2405.21066},
      archivePrefix={arXiv},
      primaryClass={cs.CV},
      url={https://arxiv.org/abs/2405.21066}, 
}

@misc{yang2024llplace3dindoorscene,
      title={LLplace: The 3D Indoor Scene Layout Generation and Editing via Large Language Model}, 
      author={Yixuan Yang and Junru Lu and Zixiang Zhao and Zhen Luo and James J. Q. Yu and Victor Sanchez and Feng Zheng},
      year={2024},
      eprint={2406.03866},
      archivePrefix={arXiv},
      primaryClass={cs.CV},
      url={https://arxiv.org/abs/2406.03866}, 
}

@misc{dong2024internlmxcomposer24khd,
      title={InternLM-XComposer2-4KHD: A Pioneering Large Vision-Language Model Handling Resolutions from 336 Pixels to 4K HD}, 
      author={Xiaoyi Dong and Pan Zhang and Yuhang Zang and Yuhang Cao and Bin Wang and Linke Ouyang and Songyang Zhang and Haodong Duan and Wenwei Zhang and Yining Li and Hang Yan and Yang Gao and Zhe Chen and Xinyue Zhang and Wei Li and Jingwen Li and Wenhai Wang and Kai Chen and Conghui He and Xingcheng Zhang and Jifeng Dai and Yu Qiao and Dahua Lin and Jiaqi Wang},
      year={2024},
      eprint={2404.06512},
      archivePrefix={arXiv},
      primaryClass={cs.CV},
      url={https://arxiv.org/abs/2404.06512}, 
}

\end{document}